\documentclass[a4paper,12pt,onecolumn]{article}

\usepackage{FAS}
\usepackage[T1]{fontenc}
\usepackage[utf8]{inputenc}
\usepackage[ngerman,english]{babel}
\usepackage{csquotes}
\usepackage{textcomp}
\usepackage{graphicx}
\usepackage{amsmath}
\usepackage[nolist]{acronym}
\usepackage[backend=biber,style=numeric-comp,sorting=none]{biblatex}
\usepackage{multirow}
\usepackage{subcaption}
\usepackage{xcolor}
\usepackage[binary-units]{siunitx}
\usepackage{float}
\usepackage{tikz}
\usepackage[normalem]{ulem}
\usetikzlibrary{arrows.meta, shapes.geometric}

\definecolor{mlorange}{HTML}{FF7F0E}
\definecolor{mlblue}{HTML}{00549f}
\definecolor{mlred}{HTML}{CC071E}
\definecolor{mlpurple}{HTML}{9467BD}
\definecolor{mlgreen}{HTML}{2CA02C}
\definecolor{mllightblue}{HTML}{9ECAE1}
\definecolor{mlgray}{HTML}{9E9E9E}
\definecolor{rwthblue75}{RGB}{64,127,183}
\definecolor{rwthred75}{RGB}{216,92,65}

\usepackage{pict2e}
\newcommand{\colorcircle}[1]{
  {\unitlength=1ex
   \begin{picture}(1.5,1)
     \linethickness{0.1ex}
     \put(0.7,0.7){\color{#1}\circle*{1.6}}
     \put(0.7,0.7){\color{black}\circle{1.6}}
   \end{picture}}%
}

\begin{document}
% Title and authors
\title{Learning-Based Behavior Planning for Automated Driving: Real-World Integration and Deployment}
\author{Jean-Pierre Busch% Prevent space before the footnote.
\thanks{Both authors are doctoral researchers at the Institute for Automotive
Engineering (ika),\\\hspace*{1.8em}RWTH Aachen University, Germany. \texttt{\{firstname.lastname\}@ika.rwth-aachen.de}}
, Guido Linden$^*$% Prevent unwanted inter-author whitespace.
, Jan Bergmann%
\thanks{Jan Bergmann contributed to this work during his master's thesis at ika\\\hspace*{1.8em}and is now with the Chair of Automotive Technology (FTM), Technical University of Munich.}
\ and Lutz Eckstein% Prevent space before the footnote.
\thanks{Lutz Eckstein is head of the Institute for Automotive Engineering (ika).}}
% Omit the date.
\date{}

% Render the title block.
\maketitle \thispagestyle{empty}

\begin{abstract}
Recent research in machine and deep learning has shown the potential of learning-based motion planning approaches to improve the driving behavior of automated vehicles, especially in complex environments. However, their complex nature and lack of transparency can hinder explainability and trustworthiness and complicate safety assurance. Motivated by these challenges, we propose a hybrid planning architecture that combines the advantages of machine learning with the verifiability and the determinism of classical approaches. Specifically, we developed a deep neural network to interpret complex traffic scenes and propose driving behavior, while an optimization-based supervision layer validates this proposal and enforces explicit drivability and safety constraints. We evaluate the learned planner's driving behavior in open-loop studies on real-world urban data, discuss system integration aspects for stable closed-loop operation, and report results from real-world deployment on our research vehicle \textit{karl.}.
\end{abstract}

\begin{keywords}
automated driving, hybrid planning architecture, behavior planning, deep learning
\end{keywords}

\begin{acronym}[MLOps]
  \acro{ATC}{Aldenhoven Testing Center}
  \acro{ADS}{automated driving system}
  \acro{ADE}{average displacement error}
  \acro{BEV}{bird's-eye view}
  \acro{DNN}{deep neural network}
  \acro{GRU}{gated recurrent unit}
  \acro{HD}{high-definition}
  \acro{ML}{machine learning}
  \acro{MLP}{multilayer perceptron}
  \acro{MLOps}{machine learning operations}
  \acro{MPC}{model predictive control}
  \acro{MCTS}{Monte Carlo Tree Search}
  \acro{OCP}{optimal control problem}
  \acro{PID}{proportional-integral-derivative}
  \acro{POMDP}{partially observable Markov decision process}
  \acro{PRM}{probabilistic roadmap}
  \acro{RRT}{rapidly-exploring random tree}
  \acro{V2X}{vehicle-to-everything}
  \acro{VRU}{vulnerable road user}
\end{acronym}

%%%%%%%%%%%%%%%%%%%%%%%%%%%%%%%%%%%%%%%%%%%%%%%%%%%%%%%%%%%%%%%%%%%%%%%%%%%%%%%%
%%%%%%%%%%%%%%%%%%%%% Introduction %%%%%%%%%%%%%%%%%%%%%%%%%%%%%%%%%%%%%%%%%%%%%
%%%%%%%%%%%%%%%%%%%%%%%%%%%%%%%%%%%%%%%%%%%%%%%%%%%%%%%%%%%%%%%%%%%%%%%%%%%%%%%%

\section{Introduction}
\label{sec:introduction}

Although companies such as Waymo~\cite{waymoMorePeopleSooner} and Zoox~\cite{zooxLasVegas} are already operating automated vehicles in U.S. cities, motion planning remains one of the central research topics for urban automated driving.
According to Donges~\cite{donges1982}, the vehicle guidance task can be divided into the subtasks of navigation, guidance, and stabilization. Nowadays, the most significant challenges are faced at the guidance level: incorporating the vehicle's environment -- in particular, other traffic participants -- into behavior and trajectory planning makes the task highly variable and considerably enlarges the relevant solution space.

Traditionally, behavior and trajectory planning algorithms are realized using search-, sampling-, or optimization-based methods, which offer deterministic behavior and allow the system to enforce clear and transparent constraints. However, in highly interactive urban traffic, purely rule-based behavior design quickly becomes difficult to scale: the long tail of corner cases, complex interactions, and the need for continuously evolving driving behavior require substantial engineering effort.

This is a key reason why data-driven planning approaches for automated vehicles are becoming increasingly important~\cite{chen2024end}. In particular, \ac{ML} models can capture implicit interaction patterns between the vehicle and its environment that are difficult to encode in rule- or optimization-based approaches. Additionally, these models can be continuously improved via \ac{MLOps} pipelines. The way in which \ac{ML} models are integrated into vehicle guidance ranges from modular system architectures that integrate \ac{ML} components for specific functions (e.g., tactical decision making) to end-to-end approaches that operate directly on raw sensor data and jointly address perception, situation understanding and planning~\cite{wang2025alpamayo, hwang2024emma}.

Regardless of whether \ac{ML} is integrated as modular components or as an end-to-end model, all approaches face the same core challenge: they must satisfy system-level requirements such as trustworthiness, explainability, accountability, drivability, traffic-rule compliance, and safety. This has also sparked standardization efforts (e.g., ISO/TS~5083:2025) that emphasize architectural measures and safety layers around \ac{ML} components~\cite{iso5083}.

Building on this motivation, this paper proposes a hybrid planning architecture that combines deep learning-based behavior planning with an optimization-based trajectory supervision, aiming to exploit the generalization capabilities of learned behavior generation while retaining deterministic, constraint-based safeguarding of the resulting trajectories. Beyond proposing the architectural concept, this work details its concrete implementation, demonstrates its integration and operation on the research vehicle \textit{karl.}~\cite{karlPaper}, and releases substantial parts of the resulting planning framework as open source software within the OpenADS\footnote{\url{https://github.com/openads-project}} ecosystem.

%%%%%%%%%%%%%%%%%%%%%%%%%%%%%%%%%%%%%%%%%%%%%%%%%%%%%%%%%%%%%%%%%%%%%%%%%%%%%%%%
%%%%%%%%%%%%%%%%%%%%% Related Work %%%%%%%%%%%%%%%%%%%%%%%%%%%%%%%%%%%%%%%%%%%%%
%%%%%%%%%%%%%%%%%%%%%%%%%%%%%%%%%%%%%%%%%%%%%%%%%%%%%%%%%%%%%%%%%%%%%%%%%%%%%%%%

\section{Related Work}

This section reviews related work on motion planning for automated vehicles, starting with classical (search-, sampling-, and optimization-based) approaches, moving on to learning-based planners, and concluding with hybrid architectures that combine \ac{ML}-driven approaches with deterministic supervision.

Rule-based motion planners have been used since the early days of automated driving, including the DARPA Grand- and Urban Challenges. For example, the winning system of the DARPA Grand Challenge 2005 used a search procedure over dynamically feasible trajectory candidates and selected solutions via cost aggregation~\cite{thrun2006stanley}. Besides search-based planning on discretized state lattices, sampling-based motion planning methods (e.g., \acp{RRT}~\cite{lavalle1998rapidly} and \ac{RRT}*~\cite{karaman2011sampling}) generate candidate trajectories by sampling the state/control space and selecting the best candidate under a cost function and constraints. For tactical decision making under uncertainty, some \acp{ADS} formulate behavior planning as a \ac{POMDP} and solve it via \ac{MCTS}, followed by trajectory optimization to ensure feasibility~\cite{komodnext}. The main advantages of these approaches are explicit constraint handling and a traceable decision process, supporting verification and safety argumentation.

On the other hand, learning-based planners aim to overcome the limited scalability of manually encoded rule sets in complex urban environments. Over the past years, a wide range of model architectures and data representations has been explored. Early and widely used approaches represent the scene as a top-down raster image and apply convolutional backbones. ChauffeurNet, for instance, learns trajectories from rasterized scene representations and uses synthetic perturbations and additional loss terms to mitigate compounding errors in closed-loop operation~\cite{bansal2018chauffeurnet}. More recent approaches use vectorized scene encodings and transformer backbones. For example, PlanT encodes a scene as a set of planar object tokens and predicts future waypoints with a transformer encoder and recurrent decoder~\cite{renz2022plant}. Vectorized representations that encode agents and map geometry as polylines and lane graphs have become a de facto standard for scene encoding~\cite{gao2020vectornet,liang2020lanegcn}. Building on them, imitation-learning planners combine attention-based scene fusion~\cite{nayakanti2023wayformer} with transformer backbones and report strong results on large-scale planning benchmarks~\cite{cheng2024plantf,cheng2024pluto}, with some approaches modeling multi-agent interaction explicitly during decoding~\cite{huang2023gameformer}.
Besides such modular approaches, end-to-end driving policies aim to map raw sensor inputs to driving behavior and may expose intermediate reasoning traces. EMMA is a representative multimodal end-to-end approach that outputs driving-task-specific predictions from camera inputs~\cite{hwang2024emma}. NVIDIA's Alpamayo is a recent large-scale foundation model that also uses camera images as input, but predicts trajectories rather than direct vehicle control commands alongside textual reasoning traces~\cite{wang2025alpamayo,nvidia2026alpamayoBlog}.

To combine the strengths of rule- and learning-based planners, recent works increasingly explore hybrid planning architectures in which learning-based behavior is validated, refined, or replaced using deterministic checks or optimization~\cite{brito2022learning,chen2022runtime,vitelli2022safetynet,gariboldi2024hybrid}. SafetyNet~\cite{vitelli2022safetynet}, for example, combines a learned planner with an explicit safety layer that evaluates feasibility, legality, and collision risk and, upon violations, switches to a non-learning-based fallback trajectory generator. In practice, the learning-based planner is based on a structured scene representation (e.g., graph-/point-based encoders with transformer-style decoding), while the fallback is realized through a trajectory generation method based on~\cite{werling2012optimal}. Gariboldi~et~al. propose a hybrid design in which a lightweight \ac{MLP} predicts a trajectory proposal that is then consistently refined by a downstream model-predictive trajectory optimization to enforce feasibility and collision avoidance~\cite{gariboldi2024hybrid}.

This line of work also connects to recent guidance on the safe integration of \ac{ML} in \acp{ADS}. For example, ISO/TS~5083:2025 provides a state-of-the-art reference for integrating \ac{ML} components into \acp{ADS} via well-defined safety layers and architectural measures~\cite{iso5083}. This work follows a similar hybrid principle by employing a learning-based behavior planner and an optimization-based supervision layer within a modular \ac{ADS}.

%%%%%%%%%%%%%%%%%%%%%%%%%%%%%%%%%%%%%%%%%%%%%%%%%%%%%%%%%%%%%%%%%%%%%%%%%%%%%%%%
%%%%%%%%%%%%%%%%%%%%% Planning Architecture %%%%%%%%%%%%%%%%%%%%%%%%%%%%%%%%%%%%
%%%%%%%%%%%%%%%%%%%%%%%%%%%%%%%%%%%%%%%%%%%%%%%%%%%%%%%%%%%%%%%%%%%%%%%%%%%%%%%%

\section{Planning Architecture}
\label{sec:architecture}
The main objective of this work is to develop a planning architecture that combines the performance potential of learning-based behavior planning with the verifiability and deterministic constraint enforcement of classical planning methods. In particular, we address the system-level challenges of learning-based approaches highlighted in Section~\ref{sec:introduction} and translate them into architectural measures that enable stable closed-loop behavior in a real-world vehicle. The architecture is designed in a modular manner with clearly separated responsibilities, allowing the learning-based behavior planner to be exchanged and updated without modifying the safety-critical supervision and fallback modules. This separation also facilitates an \ac{MLOps} workflow in which operational data can be collected, incorporated into subsequent training and validation cycles, and used for the continuous refinement of the behavior planner. In this section, the overall hybrid architecture and its interfaces are described, while Section~\ref{sec:learning-based-planning} then details the learning-based module.

\begin{figure}[htbp]
\centering
\includegraphics[width=\textwidth]{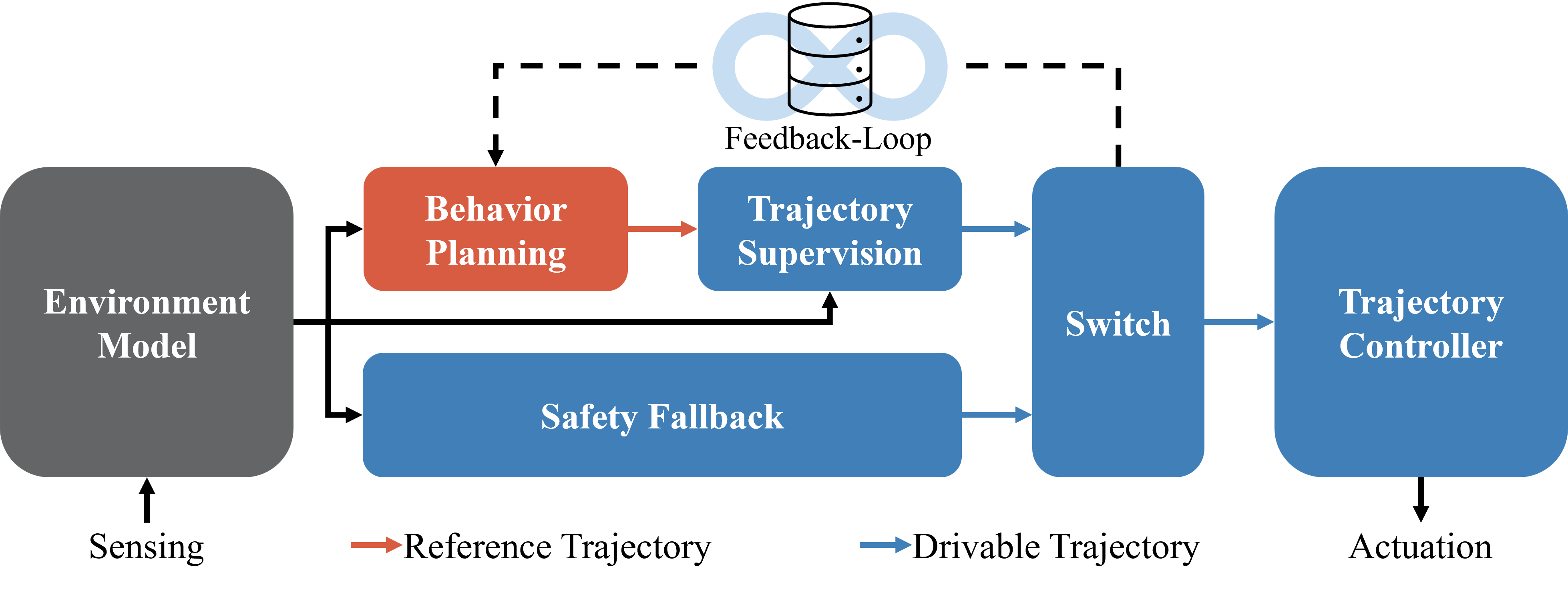}
\caption{Simplified automated driving stack focusing on the hybrid planning architecture combining learning-based\protect\colorcircle{rwthred75} and rule-based\protect\colorcircle{rwthblue75} planning modules.}
\label{fig:planning_architecture}
\end{figure}

Fig.~\ref{fig:planning_architecture} shows a simplified view of our proposed automated driving stack with a focus on the planning architecture. Upstream modules such as localization, perception, map processing, and \ac{V2X} communication provide the sensing inputs that are aggregated in an \emph{environment model}. Based on this scene representation, the learning-based \emph{behavior planning} module proposes a reference trajectory that is subsequently refined by the \emph{trajectory supervision} module under explicit drivability and safety constraints. In the context of this paper, a reference trajectory refers to a tactical decision that is not necessarily directly executable by the vehicle, but is intended to achieve the desired behavior, such as stopping behind another vehicle or performing a lane change. Parallel to the learning-based planning branch, a non-learning-based module reliably plans a drivable trajectory, acting as \emph{safety fallback}. A \emph{switch} selects the reasonable trajectory that is then tracked by the \emph{trajectory controller} and executed by the vehicle actuation.

In the following, the central rule-based modules of the architecture are described in more detail. All presented modules are implemented as C++ ROS~2 nodes and individually containerized using Docker for reproducible deployment~\cite{Busch_Enabling_the_Deployment_2024}. Substantial parts of the proposed hybrid planning architecture are released as open source software within OpenADS, where they serve as an initial planning baseline.

\subsubsection*{Trajectory Supervision}
The \emph{trajectory supervision} module safeguards the learning-based reference trajectory before it is passed to downstream modules. Its role in the architecture is to transform the reference provided by the \emph{behavior planning} module into a drivable trajectory while enforcing explicit safety, traffic-rule compliance, drivability, and vehicle-dynamic constraints.

For this purpose, we developed a trajectory optimization framework\footnote{\url{https://github.com/openads-project/trajectory_optimization}} that formulates and solves an \ac{OCP}. To ensure general drivability, a kinematic single-track model is used. The state vector~$\mathbf{x}$, control input vector~$\mathbf{u}$, and system dynamics~$f(\mathbf{x},\mathbf{u})$ are defined as follows:

\begin{equation}
\label{eq:kinematic_single_track_model}
\mathbf{x} =
\begin{bmatrix}
x \\ y \\ s \\ v \\ a \\ \psi \\ \delta
\end{bmatrix},
\quad
\mathbf{u} =
\begin{bmatrix}
j \\ \alpha
\end{bmatrix},
\quad
\dot{\mathbf{x}} = f(\mathbf{x},\mathbf{u}) =
\begin{bmatrix}
v\cos(\psi) \\
v\sin(\psi) \\
v \\
a \\
j \\
\dfrac{v}{L}\tan(\delta) \\
\alpha
\end{bmatrix}.
\end{equation}

Here, $(x,y)$~denotes the position, $s$~the distance along the trajectory, $v$,~$a$, and~$j$~the longitudinal velocity, acceleration, and jerk, respectively, $\psi$~the heading angle, $\delta$~the steering angle, $\alpha$~the steering rate, and~$L$~the wheelbase. For the discretized \ac{OCP}, the cost function~$J$~is defined as:

\begin{equation}
\label{eq:cost_function}
\begin{aligned}
J
&=
\sum_{k=0}^{N-1} \ell_k(\mathbf{x}_k,\mathbf{u}_k)
+
\ell_N(\mathbf{x}_N)
\\[0.3em]
\ell_k
&=
q_k
\left(
w_{\mathrm{lat}} \cdot e_{\mathrm{lat},k}^2
+
w_v \cdot e_{v,k}^2
+
w_{a^+} \cdot e_{a^+,k}^2
+
w_{a^-} \cdot e_{a^-,k}^2
+
w_{a_{lat}} \cdot e_{a_{lat},k}^2
\right)
\\
&\quad
+
w_{j^+} \cdot e_{j^+,k}^2
+
w_{j^-} \cdot e_{j^-,k}^2
+
w_{j_{lat}} \cdot e_{j_{lat},k}^2
+
w_{\alpha} \cdot e_{\alpha,k}^2
\\[0.3em]
\ell_N
&=
w_{\psi,N} \cdot e_{\psi,N}^2
\end{aligned}
\end{equation}

The stage cost~$\ell_k$ is evaluated at each discretization step~$k$ along the prediction horizon. The residuals $e_{\mathrm{lat}}$ and~$e_v$~penalize lateral path and velocity deviation with respect to the spatially matched reference trajectory. The terms $e_{a^+}$ and~$e_{a^-}$~distinguish between positive and negative longitudinal acceleration, while $e_{a_{\mathrm{lat}}}$~penalizes lateral acceleration. Similarly, $e_{j^+}$ and~$e_{j^-}$~distinguish between positive and negative longitudinal jerk, $e_{j_{\mathrm{lat}}}$~penalizes lateral jerk, and $e_{\alpha}$~penalizes steering-rate actuation. The coefficients $w_i$~define the relative importance of the cost terms. The factor $q_k$~denotes a dynamic weighting factor along the prediction horizon and is applied to the reference-tracking and acceleration-related terms. The terminal cost $\ell_N$~penalizes the final heading-angle deviation from the reference path.

Besides the system dynamics and cost function, the \ac{OCP} incorporates constraints on states, controls, obstacle avoidance, drivable space, and vehicle-dynamic feasibility. State and control constraints impose bounds on velocity, longitudinal acceleration, steering angle, jerk, and steering rate. Drivability constraints ensure the vehicle remains within the drivable space, while obstacle constraints guarantee collision-free trajectories with respect to dynamically predicted objects. Additional nonlinear vehicle-dynamics constraints, such as limits on absolute acceleration, are included to ensure physically feasible motion.

The optimization framework is built on acados \cite{acados2022}, which is used to formulate the nonlinear \ac{OCP} and generate solver libraries for online execution within a ROS~2 node. Through this formulation, the \emph{trajectory supervision} module transforms the reference trajectory into a drivable and collision-free trajectory with respect to the modeled constraints and predicted environment.

\subsubsection*{Safety Fallback}
In addition to the learning-based planning path, the proposed architecture includes a deterministic \emph{safety fallback}. Its role is to provide a reliable, rule-based planning branch that continuously produces a drivable trajectory. If a regular fallback trajectory cannot be generated due to invalid inputs, unavailable information, or unknown traffic situations, the module initializes a safe-stop trajectory. The fallback is therefore a complementary safety layer for degraded operation and is not intended to replace the learned \emph{behavior planning} during nominal operation.

For the fallback path, we developed a simple centerline-based planner\footnote{\url{https://github.com/openads-project/simple_planner}}. Initially, the planner constructs a reference trajectory along a given map-based centerline, including encoded speed limits as reference velocity. Based on specified traffic situations, such as approaching traffic lights or conflicts with surrounding objects, the longitudinal behavior is adapted. This is achieved by reducing the velocity profile down to a complete standstill, while the lateral reference remains tied to the centerline. The presented planner is intentionally designed as a deterministic, interpretable, and computationally efficient fallback, providing a robust and drivable trajectory in, e.g., degraded situations.

The module that determines which trajectory is to be tracked by the downstream \emph{trajectory controller} is represented by the \emph{switch} shown in Fig.~\ref{fig:planning_architecture}. In principle, it is designed to compare, evaluate, and select between multiple trajectories based on a set of arbitrary rules, for example, based on planner competence or geofences. In the presented setup, the supervised output of the learning-based planner is forwarded to the \emph{trajectory controller} by default. If the reference trajectory is of insufficient quality and the supervision module is therefore unable to ensure drivability and safety, the \emph{switch} forwards the output of the deterministic fallback path. In these cases, it is also possible to collect the necessary data for generating new training sets so that the \emph{behavior planning} can handle these scenarios in the future as well. This is a central point for the continuous improvement of the system.

\subsubsection*{Trajectory Controller}
The \emph{trajectory controller} is the final module of the presented architecture and converts the selected drivable trajectory into actuator commands for vehicle execution. It compensates tracking errors, model inaccuracies, and external disturbances while preserving the trajectory selected by the upstream planning modules as closely as possible.

To achieve this, we developed a cascaded \ac{PID}-based Ackermann trajectory controller\footnote{\url{https://github.com/openads-project/ackermann_trajectory_control}}. It is structured into a longitudinal and a lateral control path, both combining feed-forward terms from drivable trajectories provided by \emph{trajectory supervision} or \emph{safety fallback} with feedback \ac{PID} control. The required longitudinal and lateral target quantities are obtained by linear interpolation using dedicated look-ahead times. Furthermore, \ac{PID} gains are scheduled over the operating velocity range, allowing the controller behavior to be adapted to varying vehicle dynamics.

The longitudinal path tracks the target velocity by means of a velocity \ac{PID} controller. In addition, the target acceleration is used as a feed-forward contribution. The resulting acceleration command is constrained by constant longitudinal acceleration and jerk limits.

The lateral path follows a cascaded feedback structure. First, a displacement \ac{PID} controller converts the lateral trajectory deviation into a desired heading correction. A subsequent heading \ac{PID} controller combines this correction with the trajectory's target heading and generates a desired yaw rate, which is mapped to a feedback curvature command using an inverse single-track model. Feedback and feed-forward curvature terms are combined and subsequently constrained by curvature, curvature rate, and curvature acceleration limits and finally converted into Ackermann steering angle commands.

To improve overall robustness, the planning architecture follows a bi-level stabilization concept inspired by Werling et al. \cite{werling2014neues}. During regular operation, \emph{trajectory supervision} uses the interpolated state from the previously generated trajectory as the initial state for the next optimization. Whenever predefined thresholds in longitudinal or lateral tracking are exceeded, replanning from the actual vehicle state is initiated. In this way, small deviations are handled by the feedback controller, whereas larger deviations initiate a new trajectory from the actual vehicle state.

%%%%%%%%%%%%%%%%%%%%%%%%%%%%%%%%%%%%%%%%%%%%%%%%%%%%%%%%%%%%%%%%%%%%%%%%%%%%%%%%
%%%%%%%%%%%%%%%%%%%%% Learning-based Behavior Planning %%%%%%%%%%%%%%%%%%%%%%%%%
%%%%%%%%%%%%%%%%%%%%%%%%%%%%%%%%%%%%%%%%%%%%%%%%%%%%%%%%%%%%%%%%%%%%%%%%%%%%%%%%

\section{Learning-based Behavior Planning}
\label{sec:learning-based-planning}

After presenting the overall architecture and its rule-based modules, this section focuses on the developed \emph{behavior planning} module. By learning from behavioral data, it can capture interaction patterns and driving styles in dense, complex urban environments, where manually defined rules and heuristics often struggle to cover the multitude of scenarios.

\begin{figure}[H]
\centering
\includegraphics[width=\textwidth]{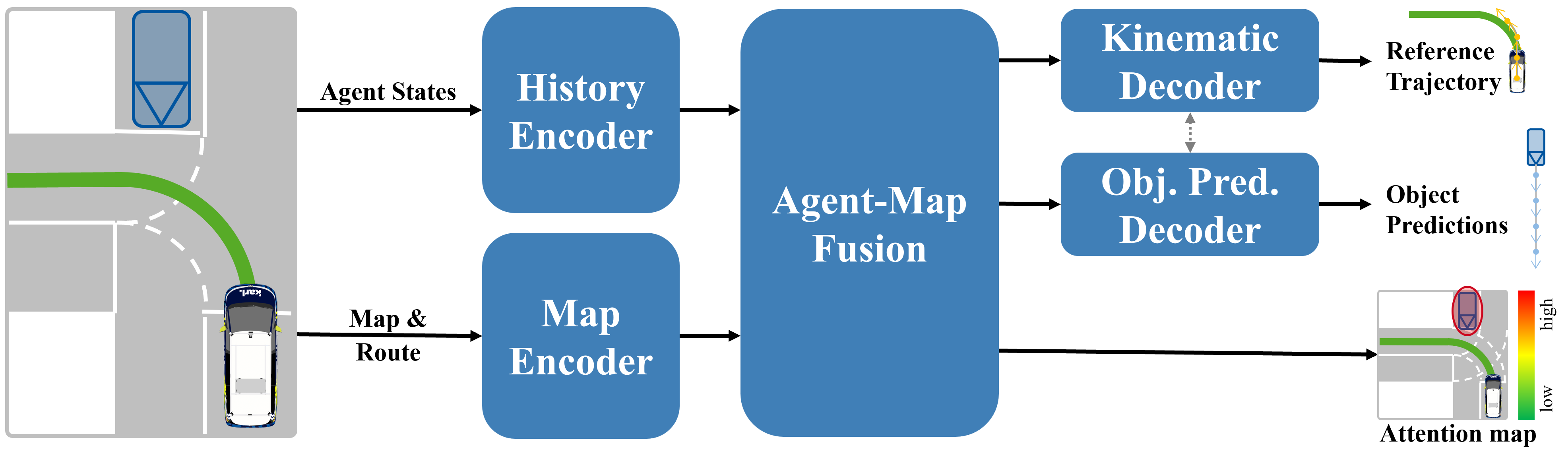}
\caption{Abstract architecture of the \ac{DNN}: Agent history and map / route information are encoded separately, fused through an attention-based fusion block, and decoded into a kinematically constrained reference trajectory. Additional outputs are object predictions, interpretable attention weights and map-grounding outputs (not illustrated).}
\label{fig:dnn_architecture}
\end{figure}

The presented behavior planner is implemented as a \ac{DNN} in PyTorch~\cite{paszke2019pytorch} and predicts a reference trajectory based on a policy learned through supervised learning from selected data. The neural network does not operate on raw sensor data, but on the processed output of the upstream environment model, including \ac{HD} map, navigation, tracked object list, and ego state information. In the following, the ego vehicle and the surrounding dynamic objects, such as other vehicles or pedestrians, are referred to as agents. The network architecture builds on a vectorized scene representation derived from the previously described data~\cite{gao2020vectornet,liang2020lanegcn} and adopts design principles from recent imitation-based transformer planners \cite{cheng2024plantf,cheng2024pluto}, including attention-based scene fusion mechanisms \cite{nayakanti2023wayformer}, adapted from motion forecasting to behavior planning and to the interfaces of the hybrid architecture. Fig.~\ref{fig:dnn_architecture} illustrates an abstract representation of the implemented network architecture while the following paragraphs detail the input and output representation, the training objective, and the used datasets. The model evaluation then follows in Section~\ref{sec:evaluation}.

\subsubsection*{Input \& Output Representation}
The input scene representation is described in an ego-centric \ac{BEV}, where all elements are expressed relative to the current ego pose and the $x$-axis is aligned with the driving direction. Each agent is represented not by a single state but by a short state history, including position, heading, velocity, and bounding box size over the last five temporal steps at a spacing of $\Delta t = \SI{0.5}{s}$, i.e.\ the past \SI{2}{s} including the current step. The object classification is kept as a separate categorical attribute.

In addition to the agent information, the map and route information are also defined in the ego-centric frame. The vectorized map comprises lane centerlines with their width and type, lane boundaries with their type and whether they may be crossed, and regulatory elements such as traffic lights, yield lines and right-of-way rules together with their current state. In addition, the map information contains the relations between centerlines, boundaries, and regulatory elements. The route is given as an ordered reference of the lanes the vehicle is meant to follow. The number of agents and map elements varies between scenes, which is handled by masking. Due to this object-level scene representation, the domain gap between data sources stays comparatively small, which makes it possible to train on various real-world or even simulation datasets.

Given this representation, the planner primarily outputs a reference trajectory over an \SI{8}{s} horizon at the same \SI{0.5}{s} spacing, represented as a sequence of positions and velocities. Rather than regressing independent waypoints, the decoder builds the trajectory stepwise. At each step it predicts a velocity and a heading change and integrates them using a simple kinematic model \cite{cui2020deepkinematic}. Per-step increments are bounded to physically reasonable ranges.

\begin{figure}[H]
\centering
\includegraphics[width=\textwidth]{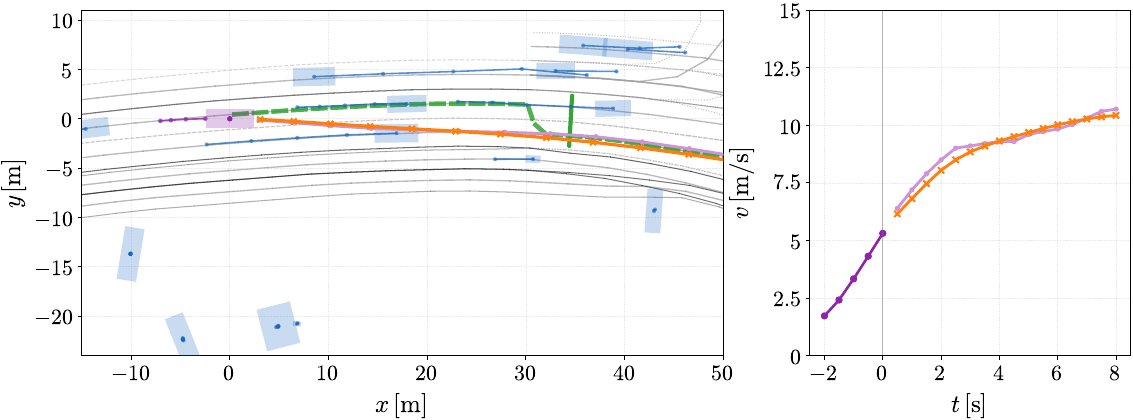}
\caption{Example of the scene representation in the ego-centric frame, shown together with the predicted reference trajectory and the corresponding velocity graph on the right.\protect\colorcircle{mlgreen}~Route, \protect\colorcircle{mlgray}~Map, \protect\colorcircle{mlpurple}~Ego (history \& future), \protect\colorcircle{mlblue}~Objects (history), \protect\colorcircle{mlorange}~Predicted~Reference~Trajectory. The solid green line represents the reference line of a green traffic light.}
\label{fig:scene_representation}
\end{figure}

Additionally, two auxiliary tasks are learned jointly with the trajectory: a short-horizon prediction of the surrounding objects, produced by a separate decoder, and a set of map-grounding quantities read from the ego decoder's per-step state, namely the lateral offset to the route and the distance to the drivable corridor. Both encourage the model to encode the interaction and the map structure. The attention weights of the fusion and the decoder remain accessible as an interpretability output. Fig.~\ref{fig:scene_representation} visualizes the scene representation and the predicted trajectory for an exemplary validation sample.

\subsubsection*{Training Objective}
As already mentioned, the planner is trained by imitation: for each recorded scene, the network is trained to predict the corresponding expert trajectory, typically originating from human driving behavior. Let $\mathbf{p}_t$, $v_t$, and $\psi_t$ be the position, velocity, and heading that the decoder produces for step $t=1,\dots,T$, and let $\hat{\mathbf{p}}_t$, $\hat v_t$, $\hat\psi_t$ be the corresponding targets, obtained from the expert trajectory. The loss combines four imitation terms with an auxiliary multi-task term,

\begin{equation}
\mathcal{L} = w_{\mathrm{p}}\,\mathcal{L}_{\mathrm{pos}} + w_{\mathrm{v}}\,\mathcal{L}_{\mathrm{vel}} + w_{\psi}\,\mathcal{L}_{\psi} + w_{\Delta\psi}\,\mathcal{L}_{\Delta\psi} + \mathcal{L}_{\mathrm{aux}},
\end{equation}
with the trajectory terms
\begin{align}
\mathcal{L}_{\mathrm{pos}} &= \frac{1}{T}\sum_{t=1}^{T}\lVert \mathbf{p}_t-\hat{\mathbf{p}}_t\rVert_1, &
\mathcal{L}_{\mathrm{vel}} &= \frac{1}{T}\sum_{t=1}^{T}\lvert v_t-\hat v_t\rvert, \\
\mathcal{L}_{\psi} &= \frac{1}{|S|}\sum_{t\in S}\lvert\mathrm{wrap}(\psi_t-\hat\psi_t)\rvert, &
\mathcal{L}_{\Delta\psi} &= \frac{1}{|S|}\sum_{t\in S}\lvert\mathrm{wrap}(\Delta\psi_t-\Delta\hat\psi_t)\rvert,
\end{align}

where $\Delta\psi_t=\mathrm{wrap}(\psi_t-\psi_{t-1})$ is the per-step heading change, $\mathrm{wrap}(\cdot)$ maps an angle to $(-\pi,\pi]$, and the heading terms are evaluated only on the moving steps $S=\{t:\hat v_t>v_{\min}\}$ with $v_{\min}=\SI{0.5}{m/s}$, where the heading is well defined.

The velocity and heading-change terms are weighted more strongly than position and absolute heading, since a faithful velocity profile and turn-in behavior are key to reproducing the reference trajectory, while the heading-change term acts directly on the decoder's control increments. The auxiliary loss, $\mathcal{L}_{\mathrm{aux}} = w_{\mathrm{a}}\,\mathcal{L}_{\mathrm{agent}} + w_{\mathrm{r}}\,\mathcal{L}_{\mathrm{route}} + w_{\mathrm{c}}\,\mathcal{L}_{\mathrm{corr}}$, consists of three lightly weighted masked regression terms supervising the auxiliary prediction heads. Their purpose is to shape the learned representation while keeping the primary focus on the imitation objective.

\subsubsection*{Datasets and Trained Models}
Building on the previously described objective, three models on different datasets are trained. The \emph{DrivIng} dataset~\cite{driving2026} serves as the primary data source for training. It comprises three recordings of the same route, approximately \SI{18}{km} long, captured at three different times (\emph{day}, \emph{dusk}, and \emph{night}) with a total duration of approximately \SI{105}{min}. The route passes through Ingolstadt in Germany, resulting in data that is characterized by urban traffic including diverse object interactions. Prior to training, object tracks were enriched with velocity estimates, and the route centerline and the relevant traffic lights were extracted from an \ac{HD} map. Traffic light states were manually annotated. Finally, samples inconsistent with the intended urban driving context or lacking relevant future information were filtered out to reduce noise.

Although the dataset is comparatively small, it enables careful curation of the training data, including detailed analysis and filtering unwanted behavior. Such a level of quality control is difficult to achieve with benchmark-scale datasets. At the same time, the limited dataset size naturally constrains the diversity of observed driving situations and therefore does not allow training a behavior planner that generalizes perfectly to arbitrary scenarios. To still compensate the reduced amount of data and further improve closed-loop performance, the data is augmented by adding noise to selected state variables of the ego vehicle and other dynamic objects, as well as applying dropout to objects that are not directly interacting with the ego vehicle.

All three models are trained on the \emph{day} and \emph{night} recordings of the \emph{DrivIng} dataset, while the \emph{dusk} recording was held out for validation. This isolates generalization across varying traffic conditions while the route geometry stays fixed. This results in \num{34438} training samples and \num{16518} validation samples, corresponding to approximately \SI{57}{min} of driving data for training and \SI{27}{min} for validation.

To study transferability to a different domain, two additional datasets were recorded on ika's test track using our research vehicle \emph{karl.} \cite{karlPaper}. The first dataset, referred to as \emph{test track}, captures driving without dynamic objects and represents the road geometry of the test track, while the second dataset (\emph{test track interaction}) includes dedicated interaction scenarios with an additional vehicle. Separate recordings were used for training and validation, yielding \num{3083}/\num{1136} training/validation samples for the \emph{test track} dataset and \num{6668}/\num{2334} samples for the \emph{test track interaction} dataset.

The first model is trained exclusively on \emph{DrivIng}, while the second and third models additionally incorporate the \emph{test track} and \emph{test track interaction} datasets, respectively. This setup allows evaluating the overall quality of the learned behavior, its transferability to a new domain, and the effect of incrementally adding domain-specific data for road-geometry adaptation and interaction behavior.

%%%%%%%%%%%%%%%%%%%%%%%%%%%%%%%%%%%%%%%%%%%%%%%%%%%%%%%%%%%%%%%%%%%%%%%%%%%%%%%%
%%%%%%%%%%%%%%%%%%%%% Evaluation and Results %%%%%%%%%%%%%%%%%%%%%%%%%%%%%%%%%%%
%%%%%%%%%%%%%%%%%%%%%%%%%%%%%%%%%%%%%%%%%%%%%%%%%%%%%%%%%%%%%%%%%%%%%%%%%%%%%%%%

\section{Evaluation and Results}
\label{sec:evaluation}
This section evaluates the proposed learning-based planning approach, from isolated model performance to full vehicle integration. First, the trained models are evaluated in isolation to assess their performance independently of the automated driving stack and to select the model used for vehicle integration. Second, the integrated model is demonstrated and qualitatively evaluated within the overall hybrid architecture.

\subsection{Model Evaluation}
\label{subsec:bp_eval}
As described in Section~\ref{sec:learning-based-planning}, the planner is trained by imitation to predict an \SI{8}{s} reference trajectory in \SI{0.5}{s} steps, together with the auxiliary outputs. Generalization is assessed in an open-loop manner by comparing the predicted reference trajectories with the recorded ego trajectories in the validation data. Tab.~\ref{tab:openloop_eval} reports the trajectory accuracy as the \ac{ADE} of position, velocity, and heading. On the \emph{DrivIng} validation split, the errors remain small: over the \SI{8}{s} horizon the position \ac{ADE} is \SI{1.83}{m}, the velocity \ac{ADE} \SI{0.56}{m/s}, and the heading \ac{ADE} \SI{2.6}{\degree}. Since the decoder determines the position by integrating the predicted velocity and heading, small but persistent errors in these quantities accumulate over the horizon, so that a velocity offset of only \SI{2}{km/h} already amounts to more than \SI{4}{m} after \SI{8}{s}. This is expected, as human drivers do not maintain a perfectly constant velocity.

\begin{table}[h]
\centering
\caption{Open-loop trajectory accuracy given as \ac{ADE} and trajectory collision ratio. All values are evaluated on distinct temporal horizons (each cell: \SI{4}{s}\,/\SI{8}{s} horizon). Rows are grouped by validation domain; within each group, the three trained models are listed based on their training data. The collision ratio is omitted for validation on \emph{test track} due to absence of objects in this data.}
\label{tab:openloop_eval}
\begin{tabular}{lcccc}
\hline Training data & Pos.\ \ac{ADE} [\si{m}] & Vel.\ \ac{ADE} [\si{m/s}] & Head.\ \ac{ADE} [\si{\degree}] & Coll.\ [\si{\percent}] \\
\hline \multicolumn{5}{l}{Validation: \emph{DrivIng}} \\ 
\hspace{1em}\emph{DrivIng} only & 0.74\,/\,1.83 & 0.37\,/\,0.56 & 1.9\,/\,2.6 & 0.3\,/\,2.3 \\
\hspace{1em}+\emph{test track} & 0.73\,/\,1.81 & 0.37\,/\,0.55 & 1.9\,/\,2.6 & 0.3\,/\,1.7 \\
\hspace{1em}+\emph{test track interact.} & 0.71\,/\,1.78 & 0.35\,/\,0.53 & 2.1\,/\,2.9 & 0.3\,/\,1.2 \\
\hline \multicolumn{5}{l}{Validation: \emph{test track}}\\
\hspace{1em}\emph{DrivIng} only & 2.26\,/\,6.96 & 1.14\,/\,1.77 & 5.0\,/\,11.0 & -- \\
\hspace{1em}+\emph{test track} & 0.66\,/\,1.31 & 0.32\,/\,0.41 & 1.7\,/\,2.0 & -- \\
\hspace{1em}+\emph{test track interact.} & 0.72\,/\,1.62 & 0.35\,/\,0.49 & 2.0\,/\,2.6 & -- \\
\hline \multicolumn{5}{l}{Validation: \emph{test track interaction}} \\
\hspace{1em}\emph{DrivIng} only & 2.54\,/\,7.83 & 1.33\,/\,2.15 & 5.3\,/\,10.0 & 6.1\,/\,26.7 \\
\hspace{1em}+\emph{test track} & 1.02\,/\,2.40 & 0.53\,/\,0.74 & 2.2\,/\,2.8 & 0.0\,/\,5.9 \\
\hspace{1em}+\emph{test track interact.} & 0.78\,/\,1.70 & 0.39\,/\,0.51 & 2.3\,/\,2.9 & 0.0\,/\,0.0 \\
\hline
\end{tabular}
\end{table}

Beyond trajectory accuracy, the learned behavior is analyzed, focusing specifically on traffic lights and their interaction with other agents: Across the entire \emph{DrivIng} validation set, the model never proposes to cross a red light, and \SI{2.3}{\percent} of the predicted \SI{8}{s} trajectories would have led to a collision, assuming that the ego vehicle followed the predicted trajectory and all other agents followed their recorded future trajectories (see Tab.~\ref{tab:openloop_eval}). Given the continuous replanning in receding-horizon operation, the full trajectory horizon is never executed. Restricting the evaluation to the first \SI{4}{s} therefore provides a more meaningful assessment in practice, with a lower corresponding collision rate of \SI{0.3}{\percent}.

\begin{figure}[htb]
\centering
\includegraphics[width=\textwidth]{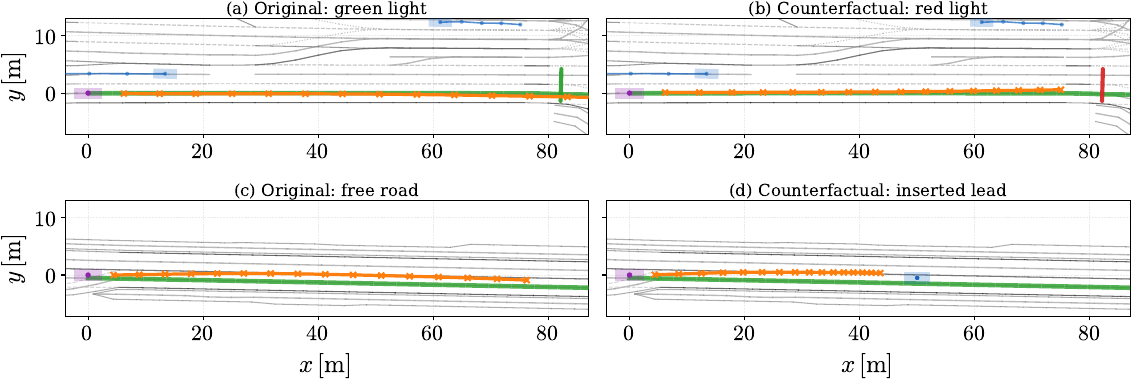}
\caption{Counterfactual validation examples. Top: changing the traffic light state from green to red causes the predicted trajectory to stop before the stop line. Bottom: inserting a stopped lead vehicle causes the planner to slow down behind the object.\protect\colorcircle{mlgreen}~Route, \protect\colorcircle{mlgray}~Map, \protect\colorcircle{mlpurple}~Ego (history), \protect\colorcircle{mlblue}~Objects (history), \protect\colorcircle{mlorange}~Predicted~Reference~Trajectory. Traffic light reference lines are colored according to their corresponding state.}
\label{fig:counterfactual}
\end{figure}

Fig.~\ref{fig:counterfactual} visualizes the learned behavior on two validation frames, each paired with a minimally modified counterfactual version. In the first pair, the originally green traffic light is switched to red. In the second pair, a stopped lead vehicle is inserted ahead of the ego vehicle. All other scene elements remain unchanged. The inserted lead vehicle is placed \SI{50}{m} ahead of the ego vehicle to ensure that collision avoidance is physically feasible. In both cases the planner adapts its prediction accordingly, stopping at the now-red traffic light and slowing down behind the inserted lead vehicle.

Transfer to a different domain is assessed on the test track recordings, whose road geometry is unseen by a model trained exclusively on \emph{DrivIng}. Zero-shot, performance is clearly degraded because the training data does not sufficiently cover the new road geometry, which shows in the much larger heading and velocity errors (see Tab.~\ref{tab:openloop_eval}; the heading \ac{ADE} rises from about \SI{2.6}{\degree} to roughly \SI{11}{\degree}), and the model collides frequently in the interaction scenarios (see Tab.~\ref{tab:openloop_eval}; \SI{26.7}{\percent} at \SI{8}{s}). When additionally trained on test track data, however, the model is able to follow the test track geometry, while the interaction behavior learned from \emph{DrivIng} transfers effectively to the new domain. The resulting model remains collision-free over the \SI{4}{s} horizon, with only a few number of residual collisions occurring over \SI{8}{s}. A small number of \emph{test track interaction} samples removes these residual collisions as well. The auxiliary tasks contribute to this transfer. Trained without them, the same \mbox{+\emph{test track}} model still collides in \SI{20.6}{\percent} of the \SI{8}{s} interaction trajectories, which the auxiliary objectives reduce to the \SI{5.9}{\percent} reported in Tab.~\ref{tab:openloop_eval}.

Conversely, incorporating the limited set of \emph{test track interaction} examples does not degrade performance on \emph{DrivIng}; instead, it yields a slight improvement, reducing the collision rate on the validation set from \SI{2.3}{\percent} to \SI{1.2}{\percent}. This result suggests that even small amounts of diverse training data can improve interaction behavior without measurable catastrophic forgetting, highlighting the potential for continuous, data-driven model improvement. For the vehicle integration described in Section~\ref{sec:integration}, the third model is used, which is trained on all available data and exhibits the most consistent generalization performance across the domains.

\subsection{Integration into a Real-World Demonstrator}
\label{sec:integration}

\begin{figure}[h]
\centering
\includegraphics[width=\textwidth,trim={0 40cm 25cm 60cm},clip]{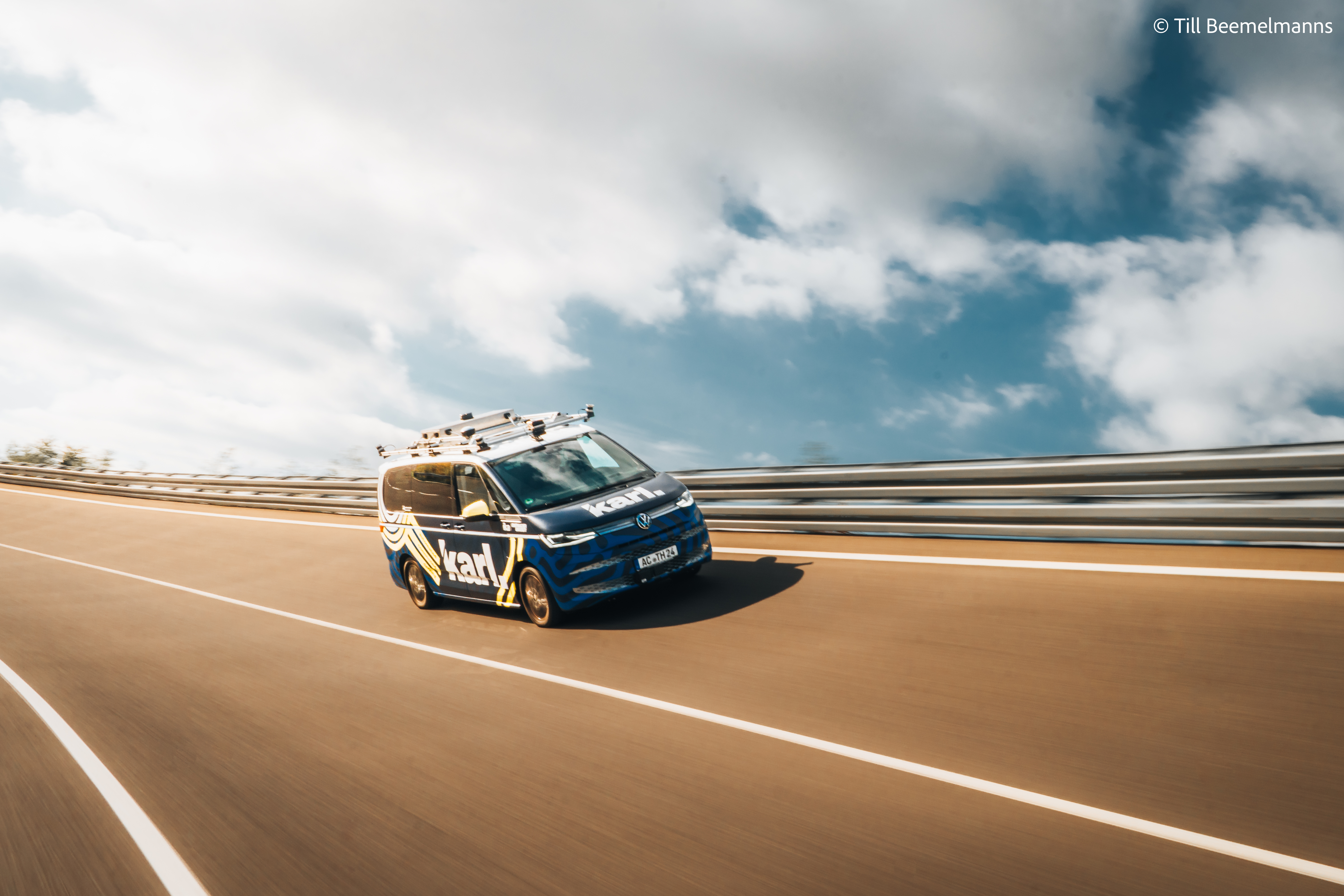}
\caption{ika's research vehicle \textit{karl.} used for the integration experiments. \protect\cite{karlPaper}}
\label{fig:karl}
\end{figure}

We deployed the proposed hybrid planning architecture on our research vehicle \textit{karl.} (cf. Fig.~\ref{fig:karl}). In contrast to purely simulation-based studies, deployment on a real-world demonstrator exposes the integrated system to realistic sensor artifacts, actuation constraints, and timing effects. It therefore enables a qualitative assessment of whether the individual modules can operate jointly under real-world conditions.
At the current stage of development, vehicle operation is restricted to a proving ground. Accordingly, all real-world experiments reported in this paper were conducted on ika's test track. Detailed descriptions of the research vehicle platform and the underlying software architecture are provided in \cite{karlPaper}.

\newpage

To evaluate the integration of the proposed planning architecture, a route was defined on the test track that covers the main behaviors considered in this work. Since the test track is not equipped with traffic lights, traffic light-related behavior is not included in the evaluation. Instead, the defined scenario focuses on lane following, lane changing, turning, and interaction with another road user. Following an initial straight segment that includes a lane change, \textit{karl.} performs two left turns. During the second turn, a challenger vehicle approaches from the right and is therefore given priority by \textit{karl.}. Subsequently, \textit{karl.} follows the challenger vehicle along the remaining route.

The proposed planning architecture has been successfully integrated into \textit{karl.}. During the conducted test runs, the \emph{behavior planning} module generated trajectory predictions at a frequency of \SI{10}{Hz} and provided references that followed the predefined route and did not cause collisions. Since the scenario requires reactions to another road user, the integration comprises not only the proposed planning modules but also the required sensor drivers and perception components. Existing software developed at the institute was used for this purpose.

\begin{figure}[ht]
\centering
\begin{subfigure}[t]{0.328\textwidth}
\centering
\includegraphics[width=\linewidth]{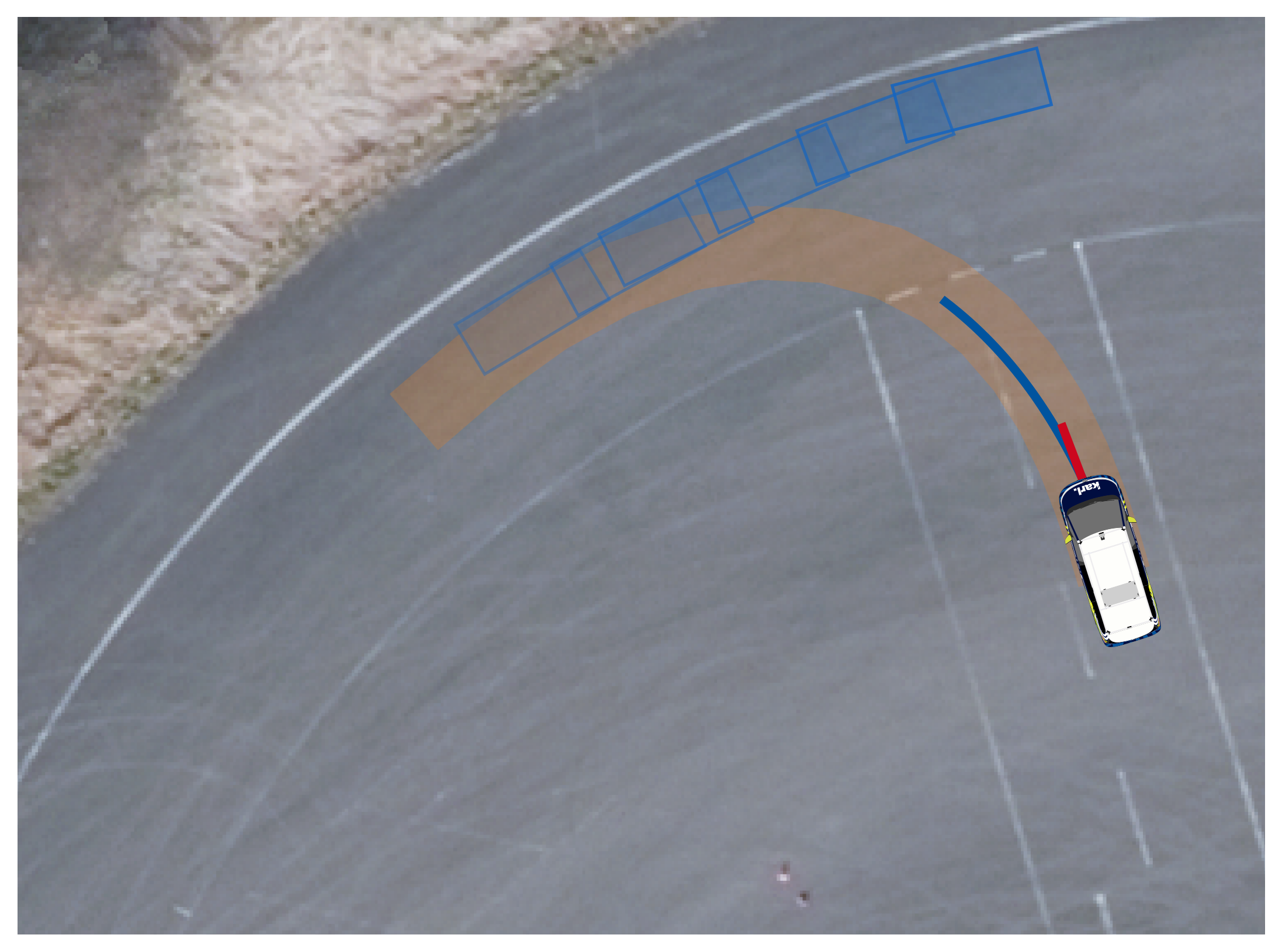}
\end{subfigure}
\hfill
\begin{subfigure}[t]{0.328\textwidth}
\centering
\includegraphics[width=\linewidth]{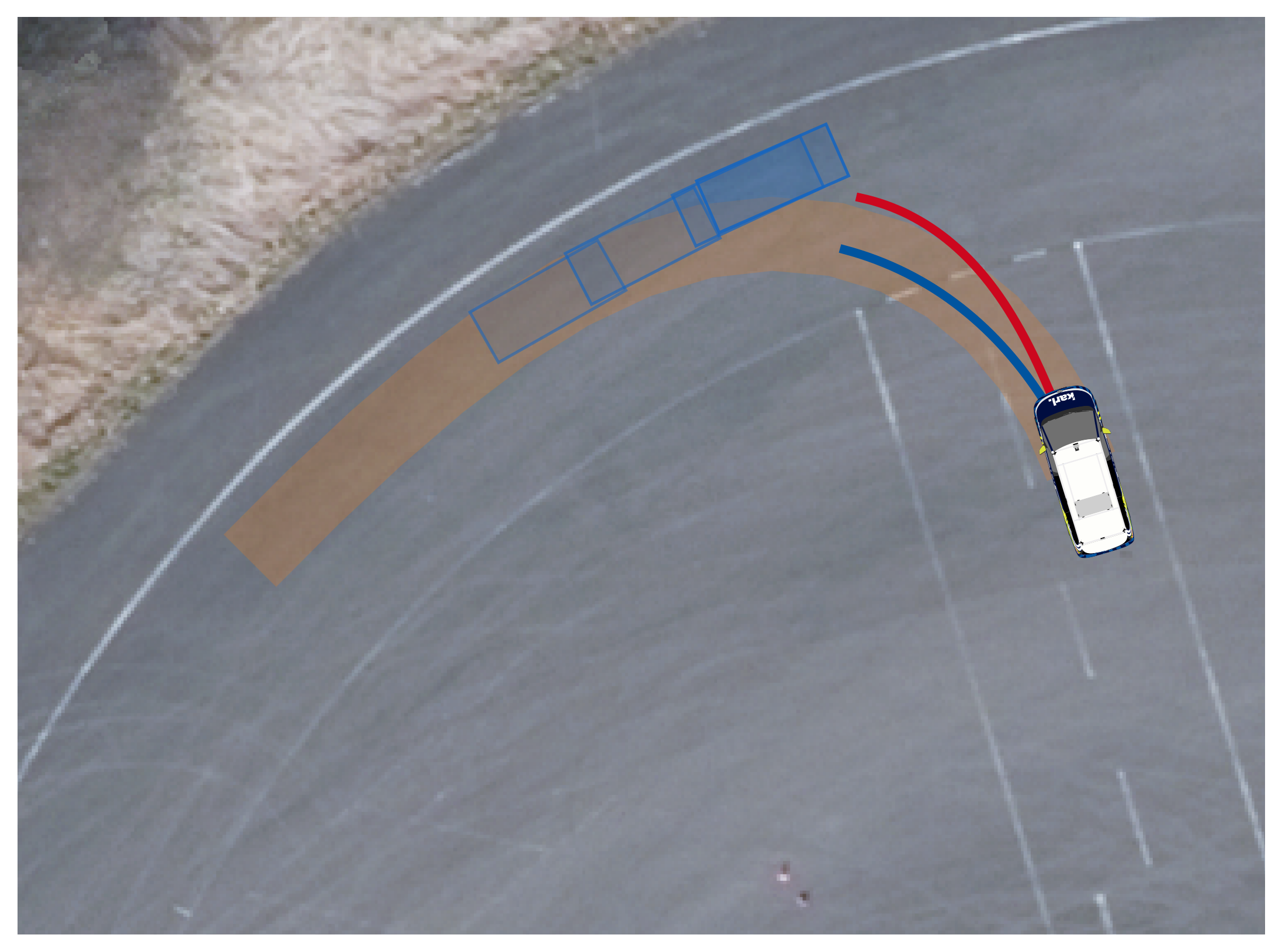}
\end{subfigure}
\hfill
\begin{subfigure}[t]{0.328\textwidth}
\centering
\includegraphics[width=\linewidth]{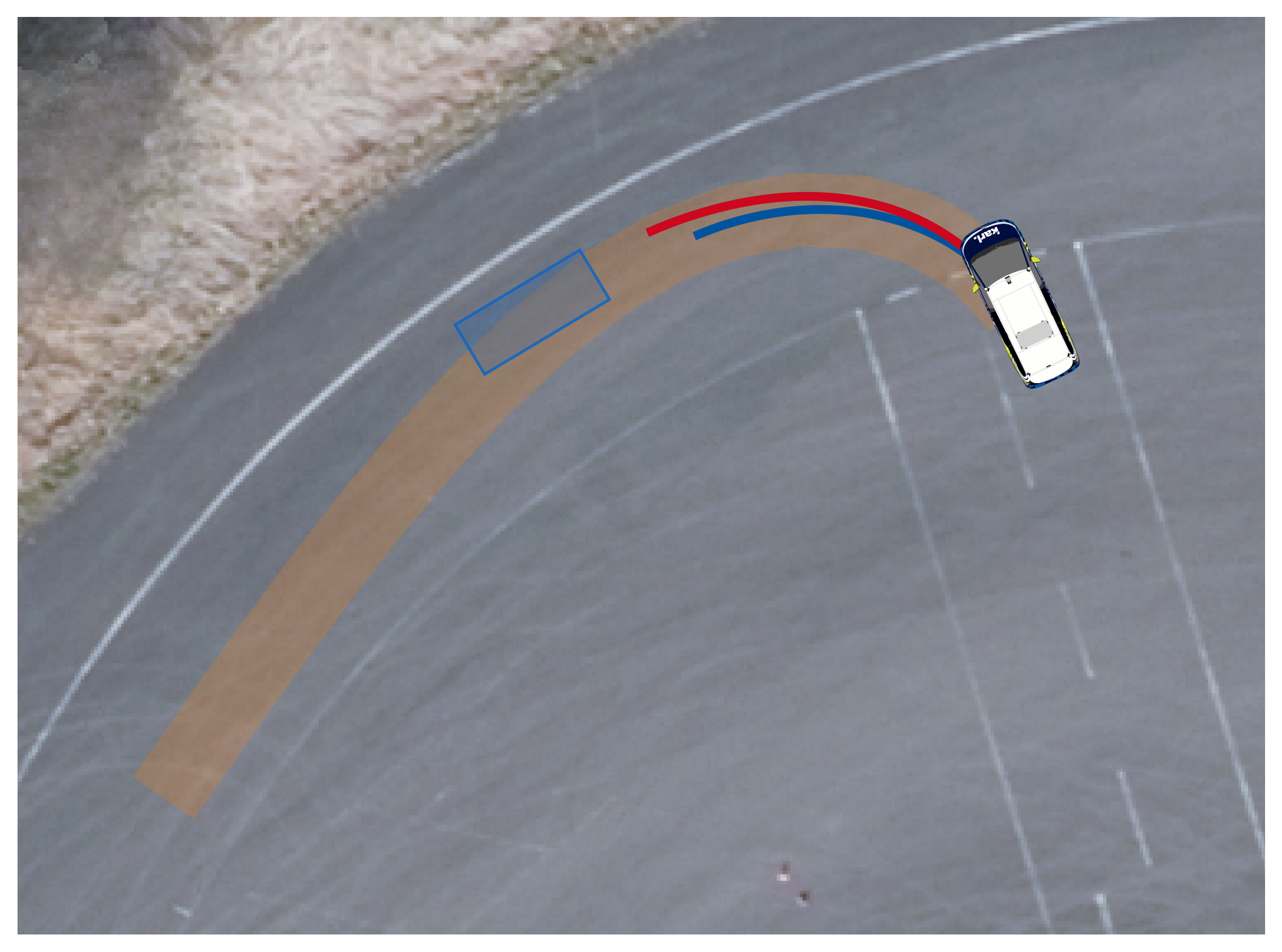}
\end{subfigure}
\caption{Consecutive \ac{BEV} snapshots of the integration experiment during the left-turn and right-of-way interaction. The orange\protect\colorcircle{mlorange} area shows the learning-based reference, the blue\protect\colorcircle{mlblue} line visualizes the trajectory obtained after \emph{trajectory supervision}, and the red\protect\colorcircle{mlred} line shows the \emph{safety fallback} trajectory computed in parallel.}
\label{fig:integration-samples}
\end{figure}

Fig.~\ref{fig:integration-samples} illustrates exemplary trajectories generated during the initial interaction with the challenger vehicle at three consecutive time steps. Each snapshot shows the learning-based reference trajectory, the supervised trajectory, and the trajectory generated by the \emph{safety fallback}. The differences in trajectory length primarily result from the different planning horizons of the corresponding modules: the \emph{trajectory supervision} and \emph{safety fallback} generate trajectories with a horizon of \SI{5}{s}, whereas the \emph{behavior planning} module predicts over a horizon of \SI{8}{s}.

Qualitatively, the snapshots indicate that the supervised trajectory largely preserves the maneuver represented by the learning-based reference while converting it into a feasible trajectory for vehicle execution. In comparison, the \emph{safety fallback} exhibits more conservative behavior, remaining closer to the lane center and decelerating more strongly when approaching the interaction area. While this behavior may provide an additional safety margin, it also results in more centerline-bound trajectories, particularly in curved road segments. The learning-based reference, by contrast, follows a smoother and less centerline-bound path through the curve, which is largely retained by the \emph{trajectory supervision}. Overall, the test track demonstration serves as a proof of concept for the real-world integration of the proposed hybrid planning architecture. It provides qualitative evidence that the individual modules can operate jointly on the research vehicle under real-world sensing, actuation, and timing conditions. Within the considered scenario, the learning-based planner provides maneuver-level reference trajectories, the supervision layer transforms these references into feasible trajectories for vehicle execution, and the conservative fallback remains available as an additional safety mechanism.

In addition to the results reported in this paper, the integrated system was demonstrated in real-vehicle operation at the \textit{autotech.agil} final event \cite{autotech-website}, where the proposed planning architecture was operated in front of invited guests at the \ac{ATC}. This public demonstration provides additional qualitative evidence that the full stack -- from perception through learning-based \emph{behavior planning} and \emph{trajectory supervision} to actuation -- can operate stably in a real-world demonstrator setting.

%%%%%%%%%%%%%%%%%%%%%%%%%%%%%%%%%%%%%%%%%%%%%%%%%%%%%%%%%%%%%%%%%%%%%%%%%%%%%%%%
%%%%%%%%%%%%%%%%%%%%% Conclusion %%%%%%%%%%%%%%%%%%%%%%%%%%%%%%%%%%%%%%%%%%%%%%%
%%%%%%%%%%%%%%%%%%%%%%%%%%%%%%%%%%%%%%%%%%%%%%%%%%%%%%%%%%%%%%%%%%%%%%%%%%%%%%%%

\section{Conclusion}
This paper presented a hybrid planning architecture for automated driving that combines learning-based behavior planning with optimization-based supervision to provide drivable and safeguarded vehicle guidance. Beyond the architectural concept, the paper provides detailed information on the implemented software modules and their integration into an operational automated driving stack. To support reproducibility and reuse, selected components of the proposed planning framework are released as open source software within the OpenADS ecosystem.

In addition, the developed \ac{ML} model and its training pipeline are described, and an open-loop evaluation on real-world urban driving data as well as findings from its integration into the research vehicle \textit{karl.} are reported. While the open-loop evaluation provides quantitative results on prediction accuracy, interaction behavior, and domain transfer, the test track experiments serve as a qualitative proof of concept for the integration of the proposed modules under real-world sensing, actuation, and timing conditions.

Overall, the results highlight the potential of hybrid architectures to couple data-driven performance with deterministic supervision and safeguarding, while additionally enabling continuous, data-driven improvement of the driving behavior. Future work will focus on a broader quantitative evaluation of the integrated system in more diverse and complex traffic scenarios. This includes evaluating the planning architecture in public road traffic once the required approval is granted.

%% Acknowledgements
\section{Acknowledgements}
We acknowledge the financial support by the German Federal Ministry of Research, Technology and Space (BMFTR) for \textit{autotech.agil} (FKZ~1IS22088A) and by the European Union's Horizon Europe Research and Innovation Programme for \textit{AIthena} (Grant Agreement No.~101076754) and \textit{AIGGREGATE} (Grant Agreement No.~101202457).

\newpage

% Bibliography
\printbibliography

\end{document}